%% file: acl_latex.tex
\pdfoutput=1

\documentclass[11pt]{article}

\usepackage[]{acl}

\usepackage{times}
\usepackage{latexsym}

\usepackage{amsmath}
\usepackage{amssymb}
\usepackage{booktabs}
\usepackage{graphicx}
\graphicspath{ {./images/} }
\usepackage{multirow}
\usepackage{cleveref}
\usepackage{diagbox}
\usepackage[utf8]{inputenc}

\definecolor{codegray}{gray}{0.9}
\definecolor{mygreen}{RGB}{126, 172, 85}
\definecolor{myred}{RGB}{147, 29, 20}

\DeclareMathOperator*{\argmax}{arg\,max}

\newcommand{\code}[1]{\colorbox{codegray}{\texttt{#1}}}

\crefname{section}{§}{§§}
\Crefname{section}{§}{§§}

\usepackage[T1]{fontenc}

\usepackage[utf8]{inputenc}

\usepackage{microtype}

\title{MetaQA: Combining Expert Agents for Multi-Skill Question Answering}

\author{Haritz Puerto$^{1}$, Gözde Gül Şahin$^{2}$, Iryna Gurevych$^{1}$\\
$^{1}$Ubiquitous Knowledge Processing Lab (UKP Lab) \\ Department of Computer Science and Hessian Center for AI (hessian.AI) \\
Technical University of Darmstadt, Germany \\
$^{2}$ Department of Computer Science,
Koç University, KUIS AI Lab \\
Istanbul, Türkiye\\ \href{https://www.ukp.tu-darmstadt.de}{https://www.ukp.tu-darmstadt.de} \&
\href{https://ai.ku.edu.tr}{https://ai.ku.edu.tr}
}

\begin{document}
\maketitle
\begin{abstract}
The recent explosion of question-answering (QA) datasets and models has increased the interest in the generalization of models across multiple domains and formats by either training on multiple datasets or combining multiple models. Despite the promising results of multi-dataset models, some domains or QA formats may require specific architectures, and thus the adaptability of these models might be limited. In addition, current approaches for combining models disregard cues such as question-answer compatibility. In this work, we propose to combine expert agents with a novel, flexible, and training-efficient architecture that considers questions, answer predictions, and answer-prediction confidence scores to select the best answer among a list of answer predictions. Through quantitative and qualitative experiments, we show that our model i) creates a collaboration between agents that outperforms previous multi-agent and multi-dataset approaches, ii) is highly data-efficient to train, and iii) can be adapted to any QA format. We release our code and a dataset of answer predictions from expert agents for 16 QA datasets to foster future research of multi-agent systems\footnote{ \href{https://github.com/UKPLab/MetaQA}{https://github.com/UKPLab/MetaQA}}.
\end{abstract}

\section{Introduction}
\input{sections/introduction}

\section{Related Work}
\label{sec:related_works}
\input{sections/related_works}

\section{Model}
\input{sections/model}

\section{Experimental Setup}
\input{sections/experimental_setup}

\section{Results and Discussions}
\input{sections/new_results}

\section{Conclusions}
\input{sections/conclusions}

\section*{Ethics Discussion}
The proposed model, MetaQA, cannot generate unfair, biased, or harmful content given that the expert agents it aggregates are fair because MetaQA does not generate content. Rather it selects from Expert Agents. The datasets we use are well-known to be safe for research purposes and do not contain any personal information or offensive content. We also comply with the licenses and intended uses of each dataset. The licenses of each dataset are shown in Appendix \ref{appendix:datasets}. We are not held responsible for errors, false or offensive content generated by the agents. MetaQA should be used at the users' discretion. Future work should address how to identify unfair or false content to avoid selecting it.

\section*{Limitations}
The main limitation of MetaQA is that when no agent has a correct answer, it returns an incorrect answer. Table \ref{table:unsolvable} describes how often this scenario occurs. In extractive datasets, without the outliers (i.e., SQuAD and DuoRC), we observe this to be 18\% on average per dataset. This percentage drops to 8.35\% in multiple-choice datasets (without BoolQ, another outlier). As for NarrativeQA and HybridQA, there are many unsolvable questions because we only use one agent for each of them and these agents have a relatively low performance.

\input{tables/no_ans_avail}

Also, if the agents do not fit in memory at the same time, it would be necessary to run them sequentially, which would increase the inference time. Yet, it might be possible to overcome this limitation because it is possible to predict in advance which agents are more likely to give a correct answer to a given question \citep{geigle:2021:arxiv, garg-moschitti-2021-will}. This would allow us to skip some agents at run-time and improve the running time dramatically in low-memory scenarios.

\section*{Acknowledgements} This work has been supported by the German Research Foundation (DFG) as part of the project UKP-SQuARE with the number GU 798/29-1 and by the German Federal Ministry of Education and Research and the Hessian Ministry of Higher Education, Research, Science and the Arts within their joint support of the National Research Center for Applied Cybersecurity ATHENE.

We thank Leonardo Ribeiro, Tim Baumgärtner, Rachneet Sachdeva, Soumya Sarkar, Neha Warikoo, and the anonymous reviewers for their insightful feedback and suggestions on a draft of this paper. We also thank Hannah Sterz for creating Adapter BART for NarrativeQA. 

\bibliography{anthology,custom}
\bibliographystyle{acl_natbib}

\clearpage

\appendix

\section{Appendix}

\subsection{Datasets}\label{appendix:datasets}
Table \ref{table:datasets} summarizes the characteristics of the datasets, contains the size of the train, validation, and test splits of each dataset, and their licenses. In the case of RACE, the authors did not provide any license but specified that it could only be used for non-commercial research purposes. In the case of CommonSenseQA and SIQA there is no license specified, but they are freely available to download. Therefore, our use of these datasets complies with their licenses and intended uses.

\subsection{Expert Agents}
\label{appendix:agents}

Table \ref{table:agents} provides the links to download the expert agents used in this work. In the case of NarrativeQA and HybridQA, we only employ one agent because of the difficulty of obtaining others. Both of these datasets use uncommon modalities (abstractive and table+text). Therefore, it is not straightforward to adapt other models to these datasets.

\subsection{Implementation}

\label{appendix:implementation}
Our model was implemented using PyTorch \citep{NEURIPS2019_9015} and HuggingFace's Transformers library \citep{wolf-etal-2020-transformers}. Both MetaQA and MultiQA were implemented using Span-BERT large (335M parameters), while UnifiedQA uses T5-base (220M parameters, the closest to the 335M of our MetaQA). The score embedder for MetaQA is implemented as a linear layer with an input size of 1 and an output size of 1024 (i.e., the hidden size of Span-BERT Large). $\alpha_1$ and $\alpha_2$ in Eq. \ref{eq:loss} are set to 0.5 and 1 respectively. The Agent Selection Networks are implemented as a linear layer with an input size of 1024 and an output size of 1. Lastly, the Answer Selection Network (AnsSel) is also implemented as a linear layer with an input size of \textit{number-of-agents} $\times$ 1025 (Span-BERT's hidden size + 1 from the output of the agent selection network). The threshold $\theta$ to determine whether a candidate answer is correct or not to create the labels to train AnsSel is set to 0.7.

MetaQA was trained for one epoch using a batch size of six, a weight decay of 0.01, a learning rate of 5e-5, and 500 warmup steps. 

All the extractive agents and MultiQA were trained using the same architecture, Span-BERT large, for two epochs and with the same hyperparameters: batch size of 16, learning rate of 3e-5, max length of 512, and doc stride of 128.

UnifiedQA was trained for two epochs using a batch size of four, a learning rate of 5e-5, and a weight decay of 0.01. It was evaluated on the dev set every 100K steps. 

Lastly, the max-voting baseline assumes that two answers are the same if the F1 score is higher than a threshold (0.9). We tuned this parameter on the dev set searching in the range $[0.5, 0.6, ...,  1.0]$. We used the implementation of HuggingFace's SQuAD F1 metric\footnote{\href{https://huggingface.co/metrics/squad}{https://huggingface.co/metrics/squad}}. In the case that two answers have the same amount of votes, we select the one with the highest confidence score given by an agent.

Any other parameter used the default value in HuggingFace's Transformers library. Each model was trained five times with different random seeds to do hypothesis testing except for UnifiedQA, which would be too expensive to compute.

We used the implementation of HuggingFace's Dataset library \citep{lhoest-etal-2021-datasets} for the evaluation using EM and F1 metrics, while for the ROGUE metric we used the official implementation\footnote{\href{https://pypi.org/project/rouge-score/}{https://pypi.org/project/rouge-score/}}.

All the experiments were conducted in a SLURM cluster where each job was assigned to different computer nodes with different CPUs and GPUs. Therefore, comparing the running time of each model is not possible.

\subsection{Adding New Agents}
Augmenting MetaQA with a new agent only requires adding one more AgSeN network and increasing the output space of the AnsSel network. Thus, it requires retraining the whole architecture (including the Transformer encoder). However, as discussed in \S \ref{sec:efficiency}, the training efficiency is one of the strengths of our system.

\input{tables/wh_stats}

\begin{table}[t]
\begin{center}
\begin{tabular}{lcc}
\toprule
Model & F1 Score \\
\midrule
MetaQA &  \textbf{73.73} \\
SpanBERT & 73.68 \\
RoBERTa & 73.15 \\
XtremeDistil & 64.16 \\
\bottomrule
\end{tabular}
\end{center}
\caption{MetaQA trained only on NewsQA agents.}
\label{table:NewsMetaQA}
\end{table} 

\subsection{MetaQA on a Single Dataset}
We conduct an additional experiment to analyze the behavior of MetaQA with multiple expert agents trained in a single dataset. We train MetaQA for three NewsQA agents: RoBERTA-base, XtremeDistil \citep{mukherjee-hassan-awadallah-2020-xtremedistil}, and SpanBERT, and evaluate it on NewsQA. As observed in Table \ref{table:NewsMetaQA}, MetaQA performs on par with the agents. However, the performance gap is smaller than in the main use case (\S \ref{sec:effectiveness}). This is attributed to the similarities between the models. These three models are all Transformers and trained on the same dataset, so it is natural that they are similar. An approach such as MetaQA excels when the agents are very different, as in Table \ref{table:main_result}, where the agents were trained on different datasets and therefore have different skills.

\input{tables/datasets_table.tex}
\input{tables/agents_table.tex}

\subsection{Wh-word Statistics}

Table \ref{table:wh_stats} shows the percentage of wh-words per dataset.

\end{document}

%% file: sections/introduction.tex
The large number of question answering (QA) datasets released in the past years has been accompanied by models specialized in them \citep{rogers2021qa, dzendzik-etal-2021-english}. These datasets and models differ by the domain (e.g., biomedical and Wikipedia), required skills (e.g., numerical and multi-hop), and format (e.g., extractive and multiple-choice).
This variety of tasks and overspecialization of the corresponding models have led the community towards developing simple unified models that can generalize across domains and formats through unifying dataset formats \citep{khashabi-etal-2020-unifiedqa}, creating models trained on multiple datasets \citep{fisch-etal-2019-mrqa, talmor-berant-2019-multiqa, khashabi-etal-2020-unifiedqa}, and designing ensemble methods for QA agents \citep{geigle:2021:arxiv}. 
All these research lines have a potential impact on end-user applications because generalization can help create robust systems and ease the implementation of QA models. For example, some chatbots are composed of skill systems, where each skill is a model trained on a specific domain \citep{miller-etal-2017-parlai, burtsev-etal-2018-deeppavlov}. More abstractly, these research lines also share a central research question: \textit{how to combine QA skills}.

\begin{figure}[t]
\centering
\includegraphics[width=\linewidth]{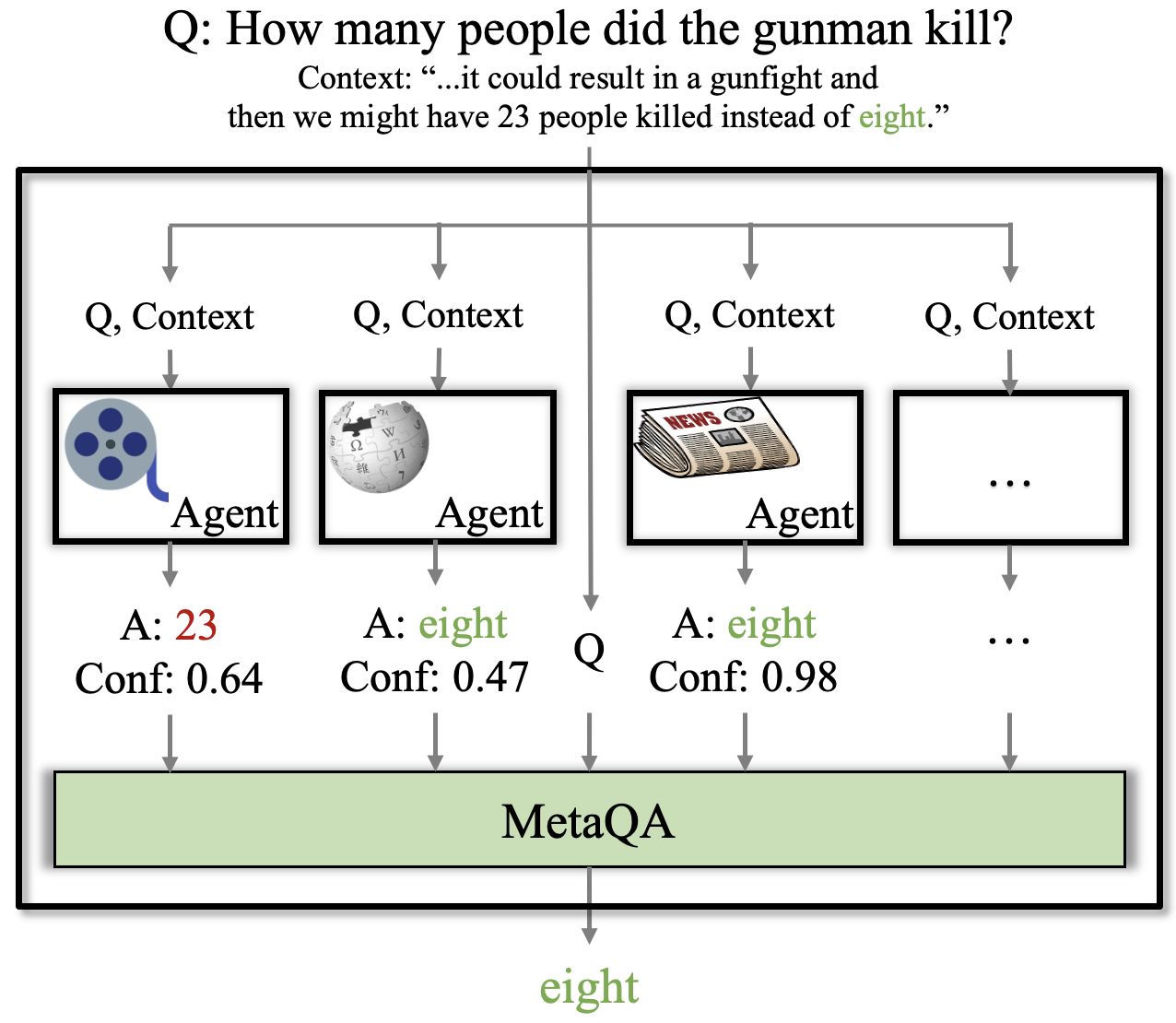}
\caption{Given a question, each expert agent provides a prediction with a confidence score and MetaQA selects the best answer. Correct answers in \textcolor{mygreen}{green}. Wrong answers in \textcolor{myred}{red}.
}
\label{fig:workflow}
\end{figure}

We argue that a \textit{one-size-fits-all} architecture may encounter some limitations in combining QA skills. For instance, \citet{raffel_t5} have observed that a single model trained on multiple tasks may underperform the same architecture trained on a single task. An alternative approach is to combine multiple expert agents. \citet{geigle:2021:arxiv} propose a model that given a question and a list of datasets, selects the dataset from which the question comes. This can be used to identify agents trained on a specific type of questions. However, despite achieving a classification accuracy greater than $90\%$, this approach underestimates high-performing models on out-of-domain questions.

To address the limitations of previous approaches, we propose a novel model, MetaQA, to combine heterogeneous expert agents (i.e., different architectures, formats, and tasks). It takes a question, and a list of \textit{candidate answers} with \textit{confidence scores} as input and selects the best answer (Figure \ref{fig:workflow}). We modify the embedding mechanism of the Transformer encoder \citep{vaswani2017attention} to embed the confidence score of each candidate answer. In addition, we use a multi-task training objective that makes the model learn two complementary tasks: \textit{selecting the best candidate answer} and \textit{identifying agents trained on the domain of the input question}.

Our approach learns to match questions with answers, an immensely easier task than the end-to-end QA of multi-dataset models. This makes MetaQA remarkably data efficient as it only uses 16\% of the training data of multi-dataset models.

We compile a list of 16 QA datasets that encompass different domains, formats, and reasoning skills to conduct experiments. Through quantitative experiments, we show that our MetaQA i) establishes a successful collaboration between agents, ii) outperforms multi-agent and multi-dataset models, iii) excels in minority domains, and iv) is highly efficient to train. Our contributions are:
\begin{itemize}
    \item A new approach for multi-skill QA that establishes a collaboration between agents.
    \item A model called MetaQA that utilizes question, answer, and confidence scores to select the best candidate answer for a given question.
    \item Extensive analyses showing the successful collaboration between agents and the training efficiency of our approach.
    \item A dataset of (\textit{QA Agents}, \textit{Questions}, and \textit{answer predictions}) triples that cover different QA formats, domains, and skills to foster future developments of multi-agent models.
    
\end{itemize}

%% file: sections/related_works.tex
Currently, there are two approaches for multi-skill QA: multi-agent and multi-dataset models.

\paragraph{Multi-agent models} 
consists of combining multiple expert agents. A well-known method is the Mixture of Experts. It requires training a set of models and combining their outputs with a gating mechanism \citep{jacobs1991adaptive}. However, this approach would require jointly training multiple agents, which can be extremely expensive, and sharing a common output space to combine the agents.
These limitations make it unfeasible to implement in our setup, where many heterogeneous agents are combined (i.e., agents with different architectures, target tasks, and output formats such as integers for multiple-choice or answer spans for span extraction). Inspired by topic classification, \citet{geigle:2021:arxiv} proposed mapping questions to QA datasets (topics) to identify agents trained on that type of questions. Although related to us, their work does not attempt to achieve any agent collaboration. Moreover, because of their \textit{topic-classification} approach, agents that are effective in out-of-domain questions are underestimated.
Lastly, \citet{friedman-etal-2021-single} average the weights of adapters \citep{houlsby2019parameter} trained on single datasets to obtain a multi-dataset model. However, their architecture is limited to span extraction.

\paragraph{Multi-dataset models} consist of training a model on various datasets to generalize it to multiple domains. \citet{talmor-berant-2019-multiqa} conduct extensive analyses of the generalization of QA models. However, they only experiment on extractive tasks and, due to their model architecture (BERT for span extraction), it is not possible to extend it to other tasks such as abstractive or visual QA. \citet{fisch-etal-2019-mrqa} created a competition on QA generalization using 18 datasets. These datasets are from very different domains, such as Wikipedia and biomedicine, among others. However, they also focus only on extractive datasets. 
Lastly, \citet{khashabi-etal-2020-unifiedqa} shows that the different QA formats can complement each other to achieve a better generalization. They use an encoder-decoder architecture and transform the questions into a common format. However, we argue that their approach is limited because some questions may require a specific skill that must be modeled in a particular manner (e.g., numerical reasoning), and this is not possible with their simple encoder-decoder.

%% file: sections/model.tex
\begin{figure}[t]
\centering
\includegraphics[width=\linewidth]{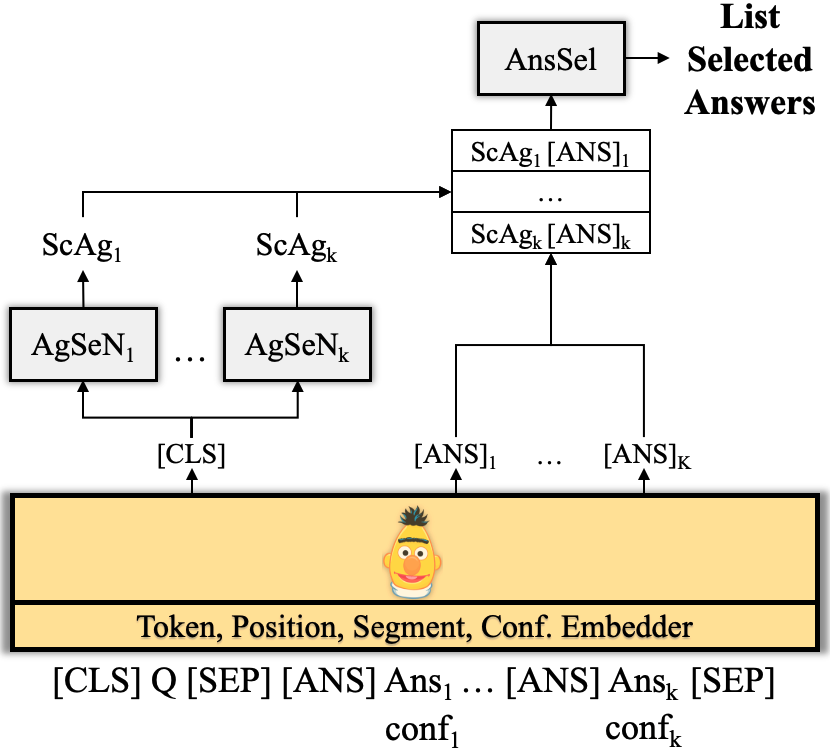}
\caption{MetaQA architecture. The Agent Selection Network, AgSeN, assigns scores to each agent based on the similarity between the agent's training set domains and the question domain. Answer Selection, AnsSel, selects the correct answers. $\text{conf}_{k}$ is the confidence score from the agent $k$ for answer $k$.}
\label{fig:Arch}
\end{figure}

We propose a new model, shown in Figure \ref{fig:Arch}, to combine $k$ QA agents $m$. Each agent $m_i$ is trained on domain $dom_i$ and predicts an answer $Ans_i$. Without loss of generalizability, we assume that each agent is trained on a different domain and each question belongs to one of these domains. We define two complementary tasks: i) \textit{a priori} agent selection (Agent Selection Networks, AgSeN, in Figure \ref{fig:Arch}) and ii) answer selection (AnsSel network in Figure \ref{fig:Arch}). The division of the problem into these two learnable tasks is vital to ensure that MetaQA considers out-of-domain agents, which can also give correct answers. More formally, MetaQA computes the probability of returning an answer $a$ as follows:
\begin{equation}
 p(a|q) =  
 \argmax_i p_{\theta_1}(a_i|m_i,q) p_{\theta_2}(m_i|q)
\end{equation}

where $p_{\theta_2}(m_i|q)$ is the \textit{a priori} probability of selecting agent $m_i$ and $p_{\theta_1}(a_i|m_i,q)$ is the probability of selecting the answer from agent $m_i$ for the question $q$. In other words, $\theta_2$ is the agent selection network, and $\theta_1$ is the answer selection network.

To achieve this, the backbone of our architecture relies on an encoder Transformer \cite{vaswani2017attention} whose input is the concatenation of the question with the candidate answers from each agent. Each answer is separated by a new token \code{[ANS]} that informs the model of the beginning of a new answer candidate.

\begin{figure}[ht]
\centering
\includegraphics[width=\linewidth]{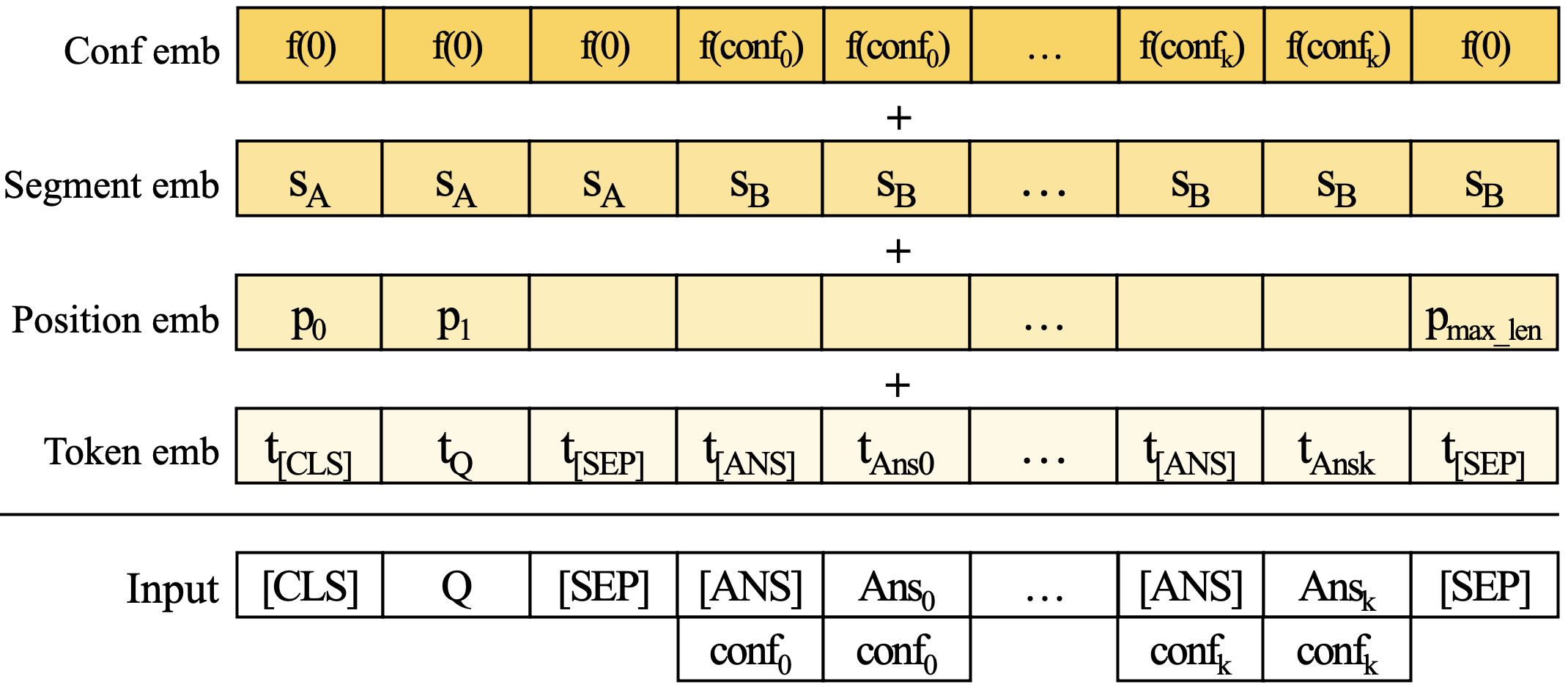}
\caption{Description of our novel embedding system including confidence scores from the agents.}
\label{fig:embedding}
\end{figure}

We devise a new embedding for the Transformer encoder to include the confidence score of the predictions of each agent (Figure \ref{fig:embedding}). While the original encoder uses the token $t_i$, position $p_i$, and segment $s_i$ embeddings, we add an agent confidence embedding $c_i$ to these three. 
\begin{equation}
    x_{i} = t_{i} + p_{i} + s_{i} + c_{i}
\end{equation}

The new $c_i$ is obtained with a feed-forward network $f$ that takes an answer confidence $\text{conf}_i$ and creates an embedding $c_i$.

\begin{equation}
    c_{i}= 
\begin{cases}
    f(\text{conf}_j),&  \text{if } i \in Idx(\text{[ANS] Ans}_j) \\
    f(0),&  \text{otherwise} 
\end{cases}
\end{equation}

where $Idx$ is a function that given a list of tokens returns their indexes in the encoder input.

We leverage two types of embeddings from the output of the encoder. The first one is the embedding of the \code{[CLS]} token. This embedding captures information about the domain of the input question. It is used as the input to $k$ independent feed-forward networks called \textit{Agent Selection Network} (AgSeN) to identify the agent with the highest likelihood of giving a correct answer. This prediction is based on the similarity between the domain of the question and the domain of the training set of the agents. More specifically, it tries to identify the agent trained on the dataset from which the question comes in a similar way as TWEAC \citep{geigle:2021:arxiv}.  The second type of embedding used is the embedding of the \code{[ANS]} tokens, which contain the cues needed to identify the correct answers to the input question. These \code{[ANS]} embeddings are concatenated with the score of each corresponding agent $ScAg_i$ and input into a final feed-forward network, called \textit{Answer Selection} (AnsSel), that selects the correct answers according to the score of their agents and the semantics of the candidate answers.

\subsection{Training}
As previously mentioned, our model learns two complementary tasks: i) \textit{a priori} agent selection and ii) answer selection. Thus, to learn these two tasks, we define the following loss function:
\begin{equation}
\label{eq:loss}
    \ell =  \frac{\alpha_1}{k}\sum_{i=0}^{k} \ell_{AgSeN_i} + \alpha_2 \ell_{AnsSel}
\end{equation}
\begin{equation}
\label{eq:AnsSel_loss}
    \ell_{AnsSel} = \frac{1}{k}\sum_{i=0}^{k} CE(\hat{Ans_i}, y_i)
\end{equation}

where $\ell_{AgSeN_i}$ is the loss of one AgSeN network and $\ell_{AnsSel}$ the loss of the AnsSel network. $\ell_{AnsSel}$ is the average of the cross-entropy loss $CE$ of each answer prediction $\hat{Ans_i} = \{0, 1\}$. Lastly, AgSeN networks use the Binary Cross Entropy.

We obtain the labels of AnsSel, $y_i$, by comparing the string prediction of each agent with the correct answer. If the F1 score is higher than a threshold, $\theta$, we consider the prediction as correct. As for $\text{AgSeN}_i$, its training label is 1 when the input question is from the training set of the $i^{th}$ agent.

%% file: sections/experimental_setup.tex
\subsection{Datasets}\label{sec:datasets}
We have collected a series of QA datasets covering different formats, domains, and reasoning skills. In particular, we use four formats: extractive, multiple-choice, abstractive, and multimodal.

For extractive, we use the MRQA 2019 shared task collection \citep{fisch-etal-2019-mrqa}, QAMR \citep{michael-etal-2018-crowdsourcing}, and DuoRC \citep{saha-etal-2018-duorc}. We include these two additional datasets to add more diversity. In detail, QAMR requires predicate-argument understanding, a skill that agents should have to solve most QA datasets. As for DuoRC, it is the only dataset in our collection on the film domain, and this allows us to study transfer learning from other domains. The multiple-choice datasets require boolean reasoning, commonsense, and passage summarization skills. Lastly, we include abstractive QA following \citep{khashabi-etal-2020-unifiedqa} and a multimodal dataset to show that our approach can solve any type of question while multi-dataset models are limited to certain formats.

Most of these datasets do not have the labels of the test set publicly available, except for RACE and NarrativeQA. Since we need to do hyperparameter tuning and hypothesis testing to compare models, we divide the public dev set into an in-house dev set and test sets following \citep{joshi-etal-2020-spanbert}. Then, we conduct hyperparameter tuning on the dev set and hypothesis testing on the test set. A summary of the datasets is available in Appendix~\ref{appendix:datasets}.

\input{tables/overall_performance_table}

\subsection{Expert Agents}
To guarantee a fair comparison with MultiQA, we have trained all the agents for extractive datasets using the same architecture as MultiQA, span-BERT, a BERT model pretrained for span extraction tasks that clearly outperforms BERT on the MRQA 2019 shared task \citep{joshi-etal-2020-spanbert}. More details on the implementation are provided in Appendix \ref{appendix:implementation}. For the remaining datasets, we use agents that are publicly available on HuggingFace or Github with a performance close to the current state of the art. A summary of them is provided in Appendix~\ref{appendix:agents}.

\subsection{Baselines}
\label{sec:baselines}
We compare our approach with three types of models: i) multi-agent systems, ii) multi-dataset models, and iii) expert agents. The first family is represented by our main baseline, TWEAC, a model that maps questions to topics (or types of questions) to identify agents trained on that type of data \citep{geigle:2021:arxiv} and the simple max-voting ensemble. The second family of models is composed of MultiQA \citep{talmor-berant-2019-multiqa} and UnifiedQA \citep{khashabi-etal-2020-unifiedqa}. MultiQA is a transformer encoder with a span-extraction layer trained on multiple extractive QA datasets. Because of this span-extraction layer, it can only solve extractive QA tasks. UnifiedQA, on the other hand, can solve any QA task that can be converted into text-to-text (i.e., extractive, abstractive, and multiple-choice) thanks to its architecture, an encoder-decoder transformer. Lastly, we include the expert agents to analyze whether MetaQA closes the gap to them compared to the baselines.

\subsection{Evaluation}
Since MetaQA may select more than one answer, we select the answer with the highest confidence score by MetaQA as the decision of the model to evaluate it.
We evaluate our model and the baselines using the official metrics of each dataset, i.e., macro-average F1 for extractive, accuracy for multiple-choice, and rouge-L for abstractive. In the particular case of DROP, the official metric is macro-average F1, and thus, we also use it. The reported results are the means and standard deviations of the models trained with five different seeds except for UnifiedQA, which would be too expensive to compute. We use a two-tailed T-Test to compare the models with a p-value of 0.05.

%% file: tables/overall_performance_table.tex
\begin{table*}[t]
\begin{center}
\begin{tabular}{lcccccc}
\toprule
\textbf{Dataset}                   & \textbf{MetaQA}     & \textbf{TWEAC} & \textbf{Exp. Agent} & \textbf{UnifiedQA} & \textbf{MultiQA} & \textbf{Voting}\\
\midrule
SQuAD           & 91.98±0.11\textdagger \phantom{\textdaggerdbl} & 89.09±0.36 & 92.92 & 90.81 & \textbf{93.14±0.18} & 90.73 \\
NewsQA          & 71.71±0.21\textdagger\phantom{\textdaggerdbl} & 66.86±0.75 & \textbf{73.68} & 65.57  & 73.59±0.60 & 66.60 \\
HotpotQA        & 79.27±0.15\textdagger\phantom{\textdaggerdbl} & 74.96±0.59 & 80.60  & 77.92 &  \textbf{81.68±0.22} & 71.71\\
SearchQA        & \textbf{81.98±0.25}\textdagger \textdaggerdbl & 80.41±0.22 & 81.04 & 81.61 &  80.45±1.82 & 68.87\\
TriviaQA-web    & \textbf{80.63±0.26}\textdagger \textdaggerdbl & 76.55±0.15 & 79.34 & 72.34 & 77.76±4.15 & 75.73\\
NQ & 81.20±0.18\textdagger \phantom{\textdaggerdbl} & 78.06±0.37 & 81.97 & 75.58 & \textbf{82.57±0.30} & 72.25\\
DuoRC           & \textbf{51.24±0.20}\textdagger \textdaggerdbl & 44.28±0.23 & 43.77 & 34.65 &  46.99±0.15 & 50.94\\
QAMR            & 83.78±0.14\textdagger \phantom{\textdaggerdbl} & 78.77±0.48 & 84.00    & 82.70 & \textbf{84.62±0.14} & 73.07\\
\midrule
BoolQ           & 73.14±0.23\textdagger\phantom{\textdaggerdbl} & 72.20±0.03  & 72.17 & \textbf{81.34} & n.a. & 73.88\\
CSQA   & \textbf{78.66±0.19}\textdagger\phantom{\textdaggerdbl} & 77.18±0.18 & 78.56 & 58.43 & n.a. & 68.41\\
HellaSWAG       & 73.19±1.01\phantom{\textdagger\textdaggerdbl} & 77.12±0.30  & \textbf{77.14} & 36.01 & n.a. & 69.33\\
RACE           & 84.71±0.05\textdagger\phantom{\textdaggerdbl} & 83.02±0.27 & \textbf{84.78} & 69.65 & n.a. & 67.30\\
SIQA            & 74.17±0.64\phantom{\textdagger\textdaggerdbl}  & 75.39±0.05 & \textbf{75.44} & 61.62 & n.a. & 70.01\\
\midrule
DROP            & 73.04±1.98\textdagger\phantom{\textdaggerdbl}  & 69.12±0.36 & \textbf{74.61} & 42.45 & n.a. & 26.18\\
NarrativeQA     & \textbf{67.19±0.00}\phantom{\textdagger\textdaggerdbl} & \textbf{67.19±0.00} & \textbf{67.19} & 57.82  & n.a. & \textbf{67.19}\\
\midrule
HybridQA        & \textbf{50.94±0.00}\phantom{\textdagger\textdaggerdbl}  & \textbf{50.94±0.00} & \textbf{50.94} & n.a & n.a & \textbf{50.94} \\

\bottomrule
\end{tabular}
\end{center}
\caption{MetaQA (ours) and the baselines on the test set of each dataset. Best results in bold. \textdagger \phantom{ } represents that MetaQA is statistically significant better than TWEAC.  \textdaggerdbl \phantom{ } represents that MetaQA is statistically significant better than MultiQA. n.a means that the system cannot model the dataset.}
\label{table:main_result}
\end{table*}

%% file: sections/new_results.tex
In this section, we answer the questions: i) is MetaQA able to combine multiple agents without undermining the performance of each one (\S \ref{sec:effectiveness}), ii) is it robust on out-of-domain scenarios? (\S\ref{sec:leave_one_out}), iii) how does agent collaboration work? (\S \ref{sec:qualit_analysis}), iv) how data-efficient is MetaQA? (\S \ref{sec:efficiency}), and v) what is the effect of each module of MetaQA? (\S \ref{sec:ablation}).

\subsection{Comparison with the Baselines}\label{sec:effectiveness}
\subsubsection{TWEAC}
MetaQA outperforms TWEAC in all datasets except HellaSWAG and SIQA, as shown in Table \ref{table:main_result}. On average, MetaQA achieves an average performance boost of 2.23 with respect to TWEAC, and more importantly, the performance boost is greater than 4 points on HotpotQA, DuoRC, NewsQA, QAMR, and TriviaQA. Particularly, there is an astonishing 6.8 points performance boost on DuoRC. 

The reason for these results is that TWEAC only aims to identify the agent trained on the domain of the question while we retrieve the best answer prediction, even if it comes from out-of-domain models. For instance, in DuoRC, MetaQA selects the in-domain agent only for 43\% of its questions, i.e., most of the questions are assigned to agents that are not trained on DuoRC. In this way, MetaQA establishes a collaboration between agents.

We also observe that the gap between MetaQA and TWEAC is more significant on extractive QA than on multiple-choice. This is expected due to our selection of multiple-choice datasets. The substantial differences in the format of these datasets limit the potential agent collaboration. For instance, BoolQ is the only boolean dataset, and therefore, it can only be used to solve boolean questions, which do not appear in the other multiple-choice datasets. Also, SIQA, a commonsense reasoning dataset, uses a short context passage while CSQA (commonsense too) does not have any context, and hence, an agent trained for CSQA cannot be used successfully on SIQA. These characteristics of the setup make the upper-bound performance of MetaQA to be the same as the expert agents. Yet, even with these limitations, MetaQA outperforms TWEAC in three of the five datasets. Also, the expert agents only significantly outperform MetaQA on 2/5 datasets. Lastly, the performance in NarrativeQA and HybridQA is the same because there is only one agent per dataset.
\input{tables/leave_one_out_table2}

\subsubsection{UnifiedQA}
MetaQA outperforms UnifiedQA by striking 10.49. In the case of extractive QA, the gap is 5.08, while in multiple-choice is 15.36. We attribute this to the limitations of UnifiedQA's architecture. For example, the performance in DROP is clearly far from our MetaQA. The reason for this is that while the expert agent used by MetaQA is designed for numerical reasoning, UnifiedQA does not have any mechanism to achieve this, and since it is designed as a general model for text-to-text generation, it cannot be augmented with special reasoning modules. The same phenomenon occurs in the multiple-choice datasets and in some minority domains in extractive QA (i.e., NewsQA and DuoRC). The only exception is in BoolQ, where UnifiedQA achieves the best results. However, this is because T5 \citep{raffel_t5}, on which UnifiedQA is trained, is already one of the SOTA models, while the agent we use has lower performance and was the only publicly available model in HuggingFace's Model Hub at the time of experimentation.

\subsubsection{MultiQA}\label{sec:multiqa}
MetaQA outperforms MultiQA by a small margin of 0.12, despite being much more flexible (i.e., compatible with any QA format instead of only extractive QA (\S \ref{sec:baselines}). Moreover, our model was trained on only 13\% of its training set, as later discussed in \S \ref{sec:efficiency}. Furthermore, we observe that MultiQA mostly outperforms expert agents on Wikipedia-based datasets (i.e., SQuAD, HotpotQA, NQ, and QAMR).
This might suggest that MultiQA is overfitted to Wikipedia due to its training on multiple datasets using Wikipedia paragraphs\footnote{MultiQA is trained on question and contexts (Wikipedia paragraphs). However, MetaQA does not have access to these paragraphs as shown in Figure \ref{fig:Arch}.} and that would explain why it struggles with other minority domains. On the other hand, MetaQA excels in minority domains where it achieves striking 4.25 points performance boost on DuoRC, 2.87 on TriviaQA-web, 1.53 on SearchQA, and overall outperforms MultiQA by an average of 2.88. These results show the superior ability of MetaQA to avoid overfitting to a specific domain.

\subsubsection{Max-Voting}
Lastly, MetaQA also outperforms max-voting by an average of 8.35. In the case of easy datasets such as SQuAD, the performance is similar because all expert agents excel in this dataset, so any approach to combine the agents would yield similar results. More interestingly, the performance of Max-Voting is clearly far from MetaQA in DROP. We attribute this to the low performance of the extractive agents on this dataset and their similar wrong answers.

\input{tables/qualitative_analysis}

\subsection{Leave-One-Out Ablation}\label{sec:leave_one_out}

In this experiment, we analyze whether the combination of expert agents can successfully solve an out-of-domain (OOD) dataset. We conduct a leave-one-out ablation test in both MetaQA and the baselines. In the case of MetaQA, we remove the expert agent of the target dataset, retrain MetaQA again without this dataset, and evaluate it on the target dataset. Similarly, we retrain TWEAC, UnifiedQA, and MultiQA without the target dataset and evaluate the model on the target dataset. Lastly, we also use the Max-Voting baseline without the agent trained on the target dataset. We trained MetaQA five times with different random seeds for each target dataset and report their average results. However, we could not do this for the other models due to their much higher computation costs.

Table \ref{table:leave_one_out} shows that OOD MetaQA outperforms OOD TWEAC in all datasets by an average of 6.31. The larger gap in OOD than in in-domain scenarios (Table \ref{table:main_result}) supports our hypothesis: the topic-classification approach of TWEAC disregards high-performing models in OOD, and our solution of establishing a collaboration between the agents is able to combine skills. 

OOD MetaQA also outperforms OOD UnifiedQA by a striking average of 8.13 points. In addition, in four datasets (TriviaQA-web, DuoRC, CommonsSenseQA, and HellaSWAG), the ablated MetaQA even outperforms the full UnifiedQA trained on those datasets. This further supports our approach of combining multiple agents, instead of datasets, in scenarios with a wide variety of domains and formats, where flexibility is key. 

In the particular case of MultiQA, as discussed in \S\ref{sec:multiqa}, half of its training sets are based on Wikipedia paragraphs. Therefore, removing a Wikipedia-based dataset such as HotpotQA does not remove Wikipedia contents from its training set\footnote{This is not the case for MetaQA because our input is only the questions, answer predictions, and confidence scores, not the Wikipedia paragraphs.}. As a consequence, this compromises the OOD setup. However, even under this pseudo-OOD setup, MultiQA only outperforms MetaQA by a slight margin of 0.61. 

Lastly, we analyze the Max Voting baseline in this scenario. Although prior works disregard this baseline, the results in Table \ref{table:leave_one_out} show that OOD Max Voting outperforms all the other baselines and has a similar performance to OOD MetaQA. Its average gain with respect to OOD MetaQA is 0.64. However, this is not the overall trend. OOD MetaQA outperforms OOD Max Voting in 5/8 extractive QA datasets by a considerable margin of 3.19. On the other hand, multiple-choice datasets, especially the difference in HellaSWAG, incline the average towards OOD Max Voting. Despite the promising claims of prior works \citep{talmor-berant-2019-multiqa, khashabi-etal-2020-unifiedqa} about OOD performance, these results suggest that aggregating a wide range of QA skills for different formats and domains in out-of-domain scenarios is still an open problem and non-neural baselines have strong results. Similar results have also been observed in retrieval methods, where non-neural baselines outperform supervised methods on OOD scenarios \citep{thakur2021beir}.

\subsection{Qualitative Analysis}
\label{sec:qualit_analysis}
We further analyze the behavior of our proposed model by inspecting its predictions. In particular, we investigate the collaboration between the agents for DuoRC, SearchQA, and TriviaQA, where this collaboration is particularly strong. 

In DuoRC, the most helpful out-of-domain (OOD) agent is NewsQA, with a chosen rate of 18.2\% in the test set. This might be due to the question types of DuoRC and NewsQA. DuoRC's questions are crowdsourced and are predominately \textit{who-questions} (42\% of the training set as shown in Appendix \ref{table:wh_stats}). NewsQA's questions are also crowdsourced and have a high proportion of \textit{who-questions} (24\%). The other datasets with a high amount of \textit{who-questions} are NQ and SearchQA. However, the questions of these two datasets are very different in style to DuoRC (i.e., real user queries and trivia from a TV show).
An example of this DuoRC-NewsQA agents collaboration is shown in the first row of Table \ref{table:qual_analysis}.

In TriviaQA-web, the second most commonly used agent is trained on DuoRC. We randomly sampled 50 QA pairs where DuoRC is the selected agent and returns the right answer. In 20\% of the cases, the question was about a movie or book plot, which indicates that our MetaQA successfully recognizes that this OOD agent is able to respond to this type of question. An example of this collaboration is shown in the second row in Table \ref{table:qual_analysis}.

In SearchQA, the most helpful OOD agent is TriviaQA (5\% chosen rate). This might be due to their similarities (Table \ref{table:datasets}). Within the pool of instances where the in-domain agent fails and the TriviaQA agent provides the right answer, we randomly analyzed 50 instances and discovered that in 84\% of the cases, the in-domain agent returns a partially correct answer, and in those cases, the OOD agent was able to identify the exact answer. This is another example of the successful agent collaboration achieved by our MetaQA. Even though the in-domain agent almost has the correct answer, MetaQA selects an OOD agent that gives a better answer, as shown in the last row on Table \ref{table:qual_analysis}.



\subsection{Efficiency of MetaQA}\label{sec:efficiency}
We trained MetaQA with bins of QA instances for each dataset and observed that the training converges with only 10K instances/per dataset (i.e., 160K instances, including all datasets). This is only 16\% of the data needed to train UnifiedQA (900K instances excluding HybridQA) and 13\% of the data needed to train MetaQA (600K of extractive QA instances). The reason for this large saving is that MetaQA only has to learn how to match questions with answers because it reuses publicly available agents. On the other hand, multi-dataset models need to learn how to solve questions (i.e., language understanding, reasoning skills, etc.), a much more complex task.

As for inference time, if all the agents fit on memory\footnote{In our hardware and with our experimental setup, all agents and MetaQA fit on our GPU memory.}, multi-datasets models and our MetaQA would have comparable running times. For example, compared to MultiQA, since our extractive agents use the same architecture as MultiQA, running the agents would take the same amount of time as running MultiQA. Then, we would need to select the answer. However, our MetaQA only takes 0.05s/question to select the best candidate answer. This makes it fast enough to not be noticeable by the users. On the other hand, if the agents do not fit in memory at the same time, it would be necessary to run them sequentially. Yet, this might not be a problem because it is possible to predict in advance which agents are more likely to give a correct answer to a given question \citep{geigle:2021:arxiv, garg-moschitti-2021-will}, which we leave as future work. This would allow us to skip some agents at run-time and improve the running time dramatically in low-memory scenarios.

\subsection{Ablation Study}
\label{sec:ablation}

Lastly, we quantitatively measure the impact of each feature of MetaQA on its overall performance. The first row of Table \ref{table:ablation} shows that removing the loss of the Agent Selection Network (AgSeN) hurts the performance of MetaQA. This manifests that our intuition of considering in-domain agents without falling into the \textit{argumentum ad verecundiam} fallacy is correct. Lastly, the second row shows that the confidence embeddings provide key information to MetaQA to select an answer. For instance, an in-domain agent could have a prediction with low confidence because it does not know the answer, while an out-of-domain agent could have the correct answer and be certain about it. 

\begin{table}[h]
\begin{center}
\begin{tabular}{lcc}
\toprule
Model & Avg. Downgrade \\
\midrule
$-\ell_{AgSeN}$ & -0.45 \\
$-$ Conf. Emb. & -0.46 \\
\bottomrule
\end{tabular}
\end{center}
\caption{Average performance loss across all datasets of each ablated model compared to the full model.}
\label{table:ablation}
\end{table}

%% file: tables/leave_one_out_table2.tex
\begin{table*}[t]
\begin{center}
\scalebox{0.66}{
\begin{tabular}{p{0.17\linewidth}|ccccccc|ccc|c|c}
\toprule
\textbf{Dataset}                        & \textbf{NewsQA} & \textbf{HotpotQA} & \textbf{SearchQA} & \textbf{TriviaQA} & \textbf{NQ}   & \textbf{DuoRC} & \textbf{QAMR}  & \textbf{CSQA}  & \textbf{HellaSWAG} & \textbf{SIQA}  & \textbf{DROP} & \textbf{$\Delta$} \\
\midrule
MetaQA     & 71.46 & 79.37 & 81.87 & 80.65 & 81.08 & 51.01 & 83.87 & 78.40  & 72.14 & 73.90  & 74.96 & -\\
UnifiedQA  & 65.57 & 77.92 & 81.61 & 72.34 & 75.58 & 34.65 & 82.70  & 58.43 & 36.01 & 61.62 & 42.45 & - \\
\midrule

OOD MetaQA & 62.26 & 69.41 & 66.59 & \underline{75.02} & 67.51 & \textbf{\underline{50.51}} & 72.20 & \underline{58.59} & \underline{52.13} & 59.28 & 22.14 & -\\

OOD TWEAC & 57.65 & 43.98 & 57.93  & 66.62  & 65.37 & 47.32 & 69.59 & 47.46  & 50.59  & 59.16 & 20.53 & -6.31\\
OOD UnifiedQA & 60.12  & 62.21    & 63.02    & 69.33        & 61.49 & 32.84 & 70.07 & 50.57 & 29.35     & 44.93 & 22.30 & -8.12\\
OOD MultiQA* & \textbf{63.36} & \textbf{69.44}   &  \textbf{67.94}  &  \textbf{76.09} & \textbf{68.52} & 49.89 & \textbf{72.53} & n.a. & n.a. & n.a. & n.a. & 0.61\\
OOD Max Voting & 63.25 &  67.59 & 61.76  &  73.81  & 68.27 & 50.48 & 68.92 & \textbf{58.94} & \textbf{64.03} & \textbf{63.22} & \textbf{22.46} & 0.64\\
\bottomrule
\end{tabular}
}
\caption{Results of leave-one-out ablation. Out-of-domain (OOD) models are trained on all the datasets except the target dataset. Best OOD results in bold. Underlined results reflect OOD MetaQA outperforming full UnifiedQA. $\Delta$ is the average performance gap to OOD MetaQA. * MultiQA uses a pseudo-OOD setup, see remarks in \S\ref{sec:leave_one_out}.}
\label{table:leave_one_out}
\end{center}
\end{table*}

%% file: tables/qualitative_analysis.tex
\begin{table*}[t]
\begin{center}
\scalebox{0.8}{
\begin{tabular}{lp{0.50\linewidth}p{0.20\linewidth}p{0.30\linewidth}}
\toprule
\textbf{Dataset} & \textbf{Question}                                       & \centering{\textbf{In-domain Agent}} & \textbf{OOD Agent}  \\\midrule
DuoRC   & Who does Rocky Balboa work for as an enforcer? & \textcolor{red}{Adrian}          & \textcolor{green!70!blue}{Tony Gazzo} (NewsQA Agent) \\
TriviaQA-web & Who played the character Mr Chips in the 2002 TV adaptation of Goodbye Mr Chips?                        & \textcolor{red}{Timothy Carroll} & \textcolor{green!70!blue}{MartinClunes} (DuoRC Agent)            \\
SearchQA     & This short story, written around 1820, contains the line "If I can but reach that bridge... I am safe" & \textcolor{red}{Legend}       & \textcolor{green!70!blue}{Legend of Sleepy Hollow} (TriviaQA Agent) \\ \bottomrule
\end{tabular}
}
\end{center}
\caption{Examples of questions where our MetaQA system disregard the in-domain agent due to their incorrect predictions (in red) and selects and an out-of-domain (OOD) agent that returns the right answer (in green).}
\label{table:qual_analysis}
\end{table*}

%% file: sections/conclusions.tex
In this work, we propose an alternative to multi-dataset models for multi-skill QA. We propose to combine expert agents to create a collaborative system for question answering (QA) called MetaQA. It considers questions, answer predictions, and confidence scores from the agents to select the best answer to a question. Through quantitative experiments, we show that our model avoids the limitations of multi-dataset models and outperforms the baselines thanks to the agent collaboration established. Additionally, since MetaQA learns to match questions with answers instead of end-to-end QA, it is highly data-efficient to train. We leave as future work: i) combining partially correct answer predictions to generate a better one, ii) adding new agents without retraining MetaQA by fixing most of the weights and only training the weights of the new Agent Selection Network, and iii) identifying \textit{a priori} agents that are likely to give an incorrect answer to skip them at run-time.

%% file: tables/no_ans_avail.tex
\begin{table}[ht]
\begin{center}
\scalebox{0.8}{
\begin{tabular}{lc}
\toprule
\textbf{Dataset}      & \textbf{\% Unsolvable} \\
\midrule
SQuAD                 & 3.92                                  \\
NewsQA                & 26.88                                 \\
HotpotQA              & 19.93                                 \\
SearchQA              & 13.97                                 \\
NQ & 19.15                                 \\
TriviaQA-web          & 12.25                                 \\
QAMR                  & 15.81                                 \\
DuoRC                 & 47.41                                 \\ \midrule
BoolQ                 & 1.47                                  \\
SIQA                  & 8.90                                   \\
HellaSWAG             & 8.90                                   \\
CSQA         & 9.00                                     \\
RACE                  & 6.61                                  \\ \midrule
DROP                  & 21.77                                 \\
NarrativeQA           & 55.71                                 \\
HybridQA              & 56.09       \\
\bottomrule
\end{tabular}
}
\end{center}
\caption{Percentage of unsolvable questions for our MetaQA with the selected agents, i.e., none of the agents can give a correct answer.}
\label{table:unsolvable}
\end{table}

%% file: tables/wh_stats.tex
\begin{table*}[t]
\begin{center}
\begin{tabular}{lccccccc}
\toprule
\textbf{Dataset} & \textbf{what} & \textbf{where} & \textbf{who} & \textbf{when} & \textbf{why} & \textbf{which} & \textbf{how}  \\ \midrule
SQuAD                 & 56.71 & 4.55  & 10.82 & 7.47  & 1.48 & 7.73  & 11.23 \\
NewsQA                & 49.52 & 8.54  & 24.46 & 5.01  & 0.11 & 3.17  & 9.19  \\
HotpotQA              & 37.98 & 4.61  & 22.99 & 2.22  & 0.05 & 29.39 & 2.76  \\
SearchQA              & 7.55  & 9.5   & 32.53 & 28.66 & 0.72 & 18.32 & 2.72  \\
NQ & 16.58 & 13.05 & 40.02 & 20.35 & 0.63 & 3.25  & 6.11  \\
TriviaQA-web          & 30.16 & 1.56  & 15.07 & 0.72  & 0.02 & 50.03 & 2.44  \\
QAMR                  & 61.75 & 5.23  & 17.92 & 4.59  & 0.66 & 3.04  & 6.82  \\
DuoRC                 & 35.16 & 9.68  & 42.32 & 2.06  & 2.44 & 1.89  & 6.45 \\\bottomrule
\end{tabular}
\end{center}
\caption{Statistics of wh-words per dataset.}
\label{table:wh_stats}
\end{table*}

%% file: tables/datasets_table.tex
\begin{table*}[t]
\scalebox{0.7}{
\begin{tabular}{lp{0.35\linewidth}p{0.50\linewidth}cccc}
\toprule
    & \textbf{Dataset} & \textbf{Characteristics}  & \textbf{Train} & \textbf{Dev} & \textbf{Test} & \textbf{License}           \\
\midrule
\parbox[t]{2mm}{\multirow{8}{*}{\rotatebox[origin=c]{90}{Extractive}}}      & SQuAD \citep{rajpurkar-etal-2016-squad}        & Crowdsourced questions on Wikipedia  & 6573  & 5253 & 5254     & MIT \\
 & NewsQA \citep{trischler-etal-2017-newsqa}       & Crowdsourced questions about News    & 74160  & 2106         & 2106 & MIT \\
 & HotpotQA \citep{yang-etal-2018-hotpotqa}     & Crowdsourced multi-hop questions on Wikipedia & 72928 & 2950 & 2951 & MIT \\
 & SearchQA \citep{dunn2017searchqa}     & Web Snippets, Trivia questions from J! Archive & 117384 & 8490 & 8490 & MIT \\
 & NQ \citep{kwiatkowski-etal-2019-natural} & Wikipedia, real user queries on Google Search & 104071 & 6418 & 6418 & MIT \\
 & TriviaQA-web \citep{joshi-etal-2017-triviaqa} & Web Snippets, crowdsorced trivia questions & 61688 & 3892 & 3893 & MIT \\
 & QAMR \citep{michael-etal-2018-crowdsourcing} & Wikipedia, predicate-argument understanding & 50615 & 18908 & 18770 & MIT \\
 & DuoRC \citep{saha-etal-2018-duorc} & Movie Plots from IMDb and Wikipedia & 58752 & 13111 & 13449 & MIT\\
\midrule                    
\parbox[t]{2mm}{\multirow{6}{*}{\rotatebox[origin=c]{90}{Multiple-Choice}}} & RACE \citep{lai-etal-2017-race} & Exams requiring passage summarization and attitude analysis  & 87866 & 4887 & 4934 & NA \\
 & CSQA \citep{talmor-etal-2019-commonsenseqa} & Web Snippets, common-sense reasoning & 9741 & 611 & 610 & NA \\
 & BoolQ \citep{clark-etal-2019-boolq} & Wikipedia, Yes/No questions & 9427 & 1635 & 1635 & CC BY-SA 3.0 \\ 
 & HellaSWAG \citep{zellers-etal-2019-hellaswag}    & Completing sentences using common sense & 39905 & 5021 & 5021 & MIT \\
 & SIQA \citep{sap-etal-2019-social} & Common sense in social interactions & 33410 & 977 & 977 & NA \\
                                    
 \midrule  
\parbox[t]{2mm}{\multirow{2}{*}{\rotatebox[origin=c]{90}{Abs.}}}     & DROP \citep{dua-etal-2019-drop} & Wikipedia, numerical reasoning & 77409 & 4767 & 4768 & CC BY-SA 4.0\\
 & NarrativeQA \citep{kocisky-etal-2018-narrativeqa} & Books, Movie Scripts & 32747 & 3461 & 10557 & Apache 2.0 \\

 \midrule  
\parbox[t]{2mm}{\multirow{2}{*}{\rotatebox[origin=c]{90}{MM}}}                       & HybridQA \citep{chen-etal-2020-hybridqa} & Wikipedia tables and paragraphs & 62682 & 1733 & 1733 & MIT \\
& & & & &\\

\bottomrule
\end{tabular}
}
\caption{Summary of the datasets used. Abs. stands for abstractive and MM for multi-modal.}
\label{table:datasets}

\end{table*}

%% file: tables/agents_table.tex
\begin{table*}[t]
\begin{center}
\scalebox{0.7}{
\begin{tabular}{lp{0.4\linewidth}cp{0.6\linewidth}}
\toprule
\textbf{\#} &\textbf{Expert Agents} &\textbf{Used for} &\textbf{Link}\\
\midrule
1& Span-BERT Large \citep{joshi-etal-2020-spanbert} for SQuAD  & all extractive + DROP & \href{https://huggingface.co/haritzpuerto/spanbert-large-cased_SQuAD}{https://huggingface.co/haritzpuerto/spanbert-large-cased\_SQuAD}       \\
2& Span-BERT Large for NewsQA   & all extractive + DROP & \href{https://huggingface.co/haritzpuerto/spanbert-large-cased_NewsQA}{https://huggingface.co/haritzpuerto/spanbert-large-cased\_NewsQA}         \\
3& Span-BERT Large for HotpotQA  & all extractive + DROP & \href{https://huggingface.co/haritzpuerto/spanbert-large-cased_HotpotQA}{https://huggingface.co/haritzpuerto/spanbert-large-cased\_HotpotQA} \\
4& Span-BERT Large for SearchQA & all extractive + DROP  & \href{https://huggingface.co/haritzpuerto/spanbert-large-cased_SearchQA}{https://huggingface.co/haritzpuerto/spanbert-large-cased\_SearchQA}  \\
5& Span-BERT Large for NQ & all extractive + DROP  & \href{https://huggingface.co/haritzpuerto/spanbert-large-cased_NaturalQuestionsShort}{https://huggingface.co/haritzpuerto/spanbert-large-cased\_NaturalQuestionsShort}   \\
6& Span-BERT Large for TriviaQA-web & all extractive + DROP & \href{https://huggingface.co/haritzpuerto/spanbert-large-cased_TriviaQA-web}{https://huggingface.co/haritzpuerto/spanbert-large-cased\_TriviaQA-web}  \\
7& Span-BERT Large for QAMR & all extractive + DROP  & \href{https://huggingface.co/haritzpuerto/spanbert-large-cased_QAMR}{https://huggingface.co/haritzpuerto/spanbert-large-cased\_QAMR}   \\
8& Span-BERT Large for DuoRC & all extractive + DROP & \href{https://huggingface.co/haritzpuerto/spanbert-large-cased_DuoRC}{https://huggingface.co/haritzpuerto/spanbert-large-cased\_DuoRC}  \\
9& RoBERTa Large \citep{liu2019roberta} for RACE  & all multiple choice  & \href{https://huggingface.co/LIAMF-USP/roberta-large-finetuned-race}{https://huggingface.co/LIAMF-USP/roberta-large-finetuned-race}         \\
10& RoBERTa Large for HellaSWAG & all multiple choice      & \href{https://huggingface.co/prajjwal1/roberta_hellaswag}{https://huggingface.co/prajjwal1/\-roberta\_hellaswag}   \\
11& RoBERTa Large for SIQA & all multiple choice & \href{https://huggingface.co/haritzpuerto/roberta_large_social_i_qa}{https://huggingface.co/haritzpuerto/roberta\_large\_social\_i\_qa}  \\
12& AlBERT xxlarge-v2 \citep{Lan2020ALBERT} for CSQA & all multiple choice   &  \href{https://huggingface.co/danlou/albert-xxlarge-v2-finetuned-csqa}{https://huggingface.co/danlou/\-albert-xxlarge-v2-finetuned-csqa}       \\
13& BERT Large-wwm \citep{devlin-etal-2019-bert} for BoolQ   & BoolQ  & \href{https://huggingface.co/lewtun/bert-large-uncased-wwm-finetuned-boolq}{https://huggingface.co/lewtun/\-bert-large-uncased-wwm-finetuned-boolq}      \\
14& TASE \citep{segal-etal-2020-simple} for DROP & DROP      & \href{https://github.com/eladsegal/tag-based-multi-span-extraction}{https://github.com/eladsegal/\-tag-based-multi-span-extraction}                \\
15& Adapter BART Large \citep{pfeiffer-etal-2020-adapterhub} for NarrativeQA & NarrativeQA & \href{https://huggingface.co/AdapterHub/narrativeqa}{https://huggingface.co/AdapterHub/narrativeqa}\\
16& Hybrider \citep{chen-etal-2020-hybridqa} for HybridQA & HybridQA & \href{https://github.com/wenhuchen/HybridQA}{https://github.com/wenhuchen/\-HybridQA} \\
\bottomrule
\end{tabular}
}
\caption{List of the expert agents, datasets in which they are used, and links to download.}
\label{table:agents}
\end{center}
\end{table*}